\documentclass[conference]{IEEEtran}
\IEEEoverridecommandlockouts
\usepackage{cite}
\usepackage{amsmath,amssymb,amsfonts}
\usepackage{algorithmic}
\usepackage{graphicx}
\usepackage{subfigure}
\usepackage{textcomp}

\usepackage{amsmath,amssymb,amsfonts}
\usepackage{xcolor}
\def\BibTeX{{\rm B\kern-.05em{\sc i\kern-.025em b}\kern-.08em
    T\kern-.1667em\lower.7ex\hbox{E}\kern-.125emX}}
\begin{document}

\title{
Augmenting Deep Learning Adaptation for Wearable Sensor Data through Combined Temporal-Frequency Image Encoding
}

\author{\IEEEauthorblockN{\textsuperscript{1} Yidong Zhu, \textsuperscript{1} Md Mahmudur Rahman, \textsuperscript{1, 2} Mohammad Arif Ul Alam}
\IEEEauthorblockA{\textsubscript{1}\textit{Computer Science, University of Massachusetts Lowell}}
\IEEEauthorblockA{\textsubscript{2}\textit{Medicine, University of Massachusetts Chan Medical School}}
}

\maketitle

\begin{abstract}
Deep learning advancements have revolutionized scalable classification in many domains including computer vision. However, when it comes to wearable-based classification and domain adaptation, existing computer vision-based deep learning architectures and pretrained models trained on thousands of labeled images for months fall short. This is primarily because wearable sensor data necessitates sensor-specific preprocessing, architectural modification, and extensive data collection. To overcome these challenges, researchers have proposed encoding of wearable temporal sensor data in images using recurrent plots. In this paper, we present a novel modified-recurrent plot-based image representation that seamlessly integrates both temporal and frequency domain information. Our approach incorporates an efficient Fourier transform-based frequency domain angular difference estimation scheme in conjunction with the existing temporal recurrent plot image. Furthermore, we employ mixup image augmentation to enhance the representation. We evaluate the proposed method using accelerometer-based activity recognition data and a pretrained ResNet model, and demonstrate its superior performance compared to existing approaches.

\end{abstract}

\begin{IEEEkeywords}
recurrent plot, image representation, frequency and temporal domain, image augmentation, activity recognition.
\end{IEEEkeywords}

\section{Introduction}
The recent advancements in deep learning techniques have revolutionized problem-solving across various domains, encompassing generative, multitask, reinforcement, active, and transfer learning \cite{i1}. These advancements have significantly improved the efficiency and scalability of classification problems \cite{i1}. However, these deep learning models heavily rely on extensive amounts of collected data and pretraining, typically conducted on powerful computers for months, predominantly dominating the fields of computer vision and natural language processing. To extend the applicability of these models to wearable sensor data, several challenges need to be addressed, including preprocessing, artifact removal, noise reduction, and careful modification of advanced deep learning techniques. Additionally, substantial efforts are required for data collection to facilitate the pretraining process. Consequently, existing pretrained computer vision-based models, such as RestNet and AlexNET, are rendered ineffective in the context of wearable sensors.

To bridge this gap, researchers have proposed converting wearable sensor data into image representations, predominantly using Recurrent Plots in the temporal domain \cite{reference1, reference3}. In this paper, we introduce a novel modified-recurrent plot-based image representation for wearable sensor data that incorporates both temporal and frequency domain information. Firstly, we design an efficient Fourier transform-based frequency domain angular difference estimation scheme for the recurrent plot of wearable sensor readings. Building upon this, we employ an image augmentation technique called mixup to combine the temporal and frequency domain images, resulting in a comprehensive representation. Our {\bf key contributions:}

\begin{figure}[!htb]
\begin{center}
 \includegraphics[width=\linewidth]{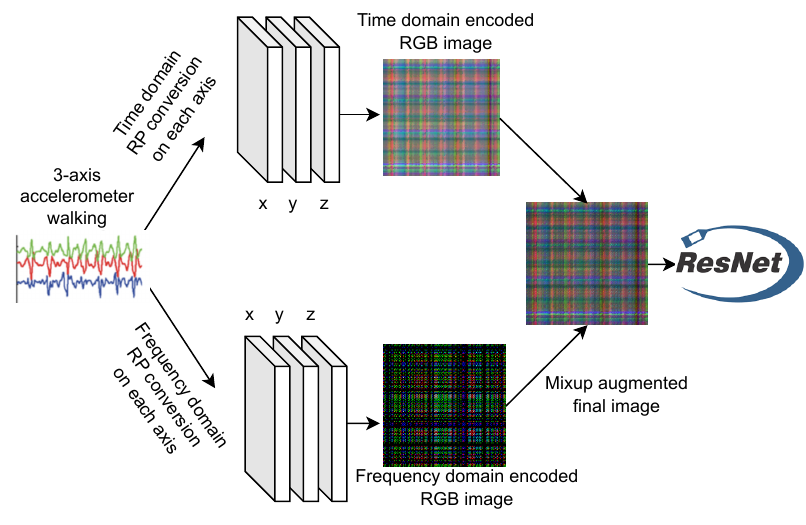}
 \caption{The schematic diagram of proposed work}
 \label{fig:overview}
\end{center}
\end{figure}
\begin{itemize}
    \item In order to construct frequency-domain modified recurrence plots, we initially examine the phasic difference within a single channel of wearable sensor data. Next, we compute the angular difference between two data points, and subsequently represent the tri-channel signals as RGB images.

    \item Using temporal images derived from Modified Recurrence Plots \cite{reference1}, we applied the MixUp augmentation technique to generate new images that encompass comprehensive information from both the time-domain and frequency-domain.

    \item  Lastly, we assessed the effectiveness of our approach by conducting evaluations on publicly available wearable 3-axis accelerometer data for activity recognition. By utilizing a pretrained ResNet model, our method showcased superior performance, surpassing the capabilities of existing techniques.
\end{itemize}

\section{Related Works}

Activity recognition is vital in wearable technology, with applications in health monitoring, medicine, psychology, and security. Previous studies used various sensors (e.g., accelerometers, magnetometers, gyroscopes) in cameras, smartphones, and watches. Algorithms like Random Forest \cite{rf}, Support Vector Machines \cite{svm}, CNNs \cite{cnn}, RNNs \cite{lstmfcn}, and LSTMs extract patterns from sensor data for accurate recognition. However, existing approaches require extensive preprocessing and hinder scalability. To address this, we propose a novel method: converting wearable sensor data to recurrence plot images and employing a pre-trained ResNet for improved scalability and performance.

Several studies have suggested transforming time series recognition problems into image classification tasks (\cite{reference1, reference2}). Gramian Angular Field (GAF) and Markov Transition Field (MTF) have been introduced for single-channel time series data (Reference 1). GAF represents the time series data in a polar coordinate system, with values fluctuating among different angular points on surrounding circles as time progresses. MTF expands transition probabilities on the magnitude axis into a matrix, considering temporal positions. Activity recognition data from wearable sensors is commonly treated as time series datasets, and classification tasks often rely on time-domain methods. Two primary approaches exist: one involves heuristic handcrafted features, while the other focuses on the time-domain shape of signal instances. The former employs features such as statistics, frequency or wavelet transform, and energy, which are then used as inputs for models like support vector machines (SVM), random forests (RF), and hidden Markov models (HMMs). However, these methods heavily rely on feature extraction quality and domain-specific knowledge \cite{reference4, reference5, reference6}. The latter approach often employs Dynamic Time Warping (DTW) to measure signal similarity, often combined with a k-nearest neighbor (k-NN) framework to improve performance \cite{reference7, reference8}. However, DTW can be slow with large datasets, despite attempts to speed up the process.

All these studies solely consider time domain information and do not incorporate frequency domain information. Their accuracy remains stagnant at 93\%. In contrast to these approaches, we propose the integration of time domain and frequency domain information using MixUp augmentation. This allows for the utilization of both domain information in classification tasks.

\section{Frequency Domain Information Encoding in Image}
\subsection{Recurrent Plot}
The Recurrence Plot (RP) is a visualization tool used to study complex dynamic systems \cite{reference25}. It represents nonlinear data points on phase space trajectories, depicting small-scale features like dots and lines, as well as large-scale textures such as homogeneity, periodicity, drift, and disruption. The RP is expressed as a matrix $\mathbf{R}$, calculated from a trajectory data sample $\mathbf{x}$, where each element $R_{i,j}(\varepsilon)$ represents the L2 norm of the difference between data points $\mathbf{x}_i$ and $\mathbf{x}_j$. To exploit correlation information, the RP is used to encode 3-axis signals as RGB channels of images. States in phase space can be represented by \(s_j = (x_j, x_{j+1})\), where \(s_j \in \mathbb{R}^2\). The recurrence plot (RP) can be formulated by a recurrence matrix \(R\), where \(R \in \mathbb{R}^{(N-1) \times (N-1)}\), whose element is the L2 norm of a state difference vector. It has the following formulation:
\begin{equation}
\label{eq:rp_matrix}
    R_{m,n} = ||s_i - s_j||
\end{equation}

\subsection{Encoding Accelerometer Signals as Images Using
Modified Recurrence Plot on Frequency Domain}
The recurrence matrix is symmetric with respect to the zero main diagonal. However, this symmetry will confuse the tendency of signals. To resolve this, researchers proposed a modified recurrent plot method for temporal domain \cite{reference1}. It proposed to calculate angle between a base vector and the temporal state difference vector \(s_m - s_n\) to identify the sign of the recurrent plot of Equation \ref{eq:rp_matrix} as follows.
\begin{equation}
\label{eq:temporal_matrix}
    R_{m,n} = sign(m,n) ||s_i - s_j||
\end{equation}
However, we further improve this method by incorporating frequency domain information in the recurrent plot. In this regard, first, we hypothesize, for frequency phase whose tendency is uphill,
their state difference vector falls in the first quadrant of
the Cartesian coordinate system, while for those in a downhill tendency, the state difference vector falls in the third quadrant.
By following this observation, we first calculate the Fourier transform of two temporal phases within their time-window resulting in the complex-valued frequency spectra. Then, we compute the phase of each frequency component noted as $p_i$ and $p_j$ corresponding to temporal phase $s_i$ and $s_j$ respectively. Now, we use the angle between a base vector $v$ and the phase difference vector to distinguish different gradient directions. For example, if the angle between the base vector $v = [1, 1]$ and the positive direction of x-axis is $\frac{1}{4\pi}$, then all vectors with an angle bigger than $\frac{3}{4\pi}$ to $v$ are in the third quadrant. Mathematically, a sign function is used whose formulation is given by
\begin{equation}
\text{sign}(m, n) = \begin{cases}
-1, & \text{if } \frac{(p^i-p_j).v}{||p^i-p^j||.||v||} <    cos(\frac{3}{4\pi}) \\
1, & \text{otherwise}
\end{cases}
\end{equation}
where $v = [1, 1]$. Thus the modified recurrent plot for frequency domain is
\begin{equation}
\label{eq:frequency_matrix}
    R_{m,n} = sign(m,n) ||p_i - p_j||
\end{equation}

We use Equation \ref{eq:frequency_matrix} for each of our target sensor (accelerometer) channel (3-axis) to transform into three recurrent plot images. We combine these 3 images into a single matrix $M$ ($M \in \mathbb{R}^{(N-1) \times (N-1) \times 3}$). Then we normalize this matrix and encode in RGB image.

\subsection{Mixup Augmentation of Temporal and Frequency Domain RP Plot of Wearables}

Mixup image augmentation is a technique that combines pairs of images to create new augmented images \cite{mixup}. Mathematically, given two input images $x_1$ and $x_2$, mixup generates a new augmented image $x_a$ as follows:
\begin{equation}
    x_a = \lambda \cdot x_1 + (1 - \lambda) \cdot x_2
\end{equation}
Note that the $\lambda$ values are values with the [0, 1] range and are sampled from the Beta distribution. We utilize Equation \ref{eq:temporal_matrix} and Equation \ref{eq:frequency_matrix} to generate temporal and frequency domain recurrent plot of multi-channel wearable sensor generated images into a single one.

\section{Experimental Evaluation}
\subsection{Dataset}
Two distinct datasets were employed for conducting the experiments. The first dataset, known as Activities of Daily Living (ADL), was obtained from the UCI Machine Learning Repository \cite{reference2} and is widely accessible. The second dataset, named ASTRI, was originally provided by the Hong Kong Applied Science and Technology Research Institute (ASTRI).

\begin{itemize}
    \item ADL Dataset: This dataset comprises tagged wrist-worn accelerometer data collected from 16 volunteers. The data was recorded using a tri-axial accelerometer with a sampling rate of 32 Hz. It encompasses 14 different daily activities; however, for this experiment, only 7 activities were utilized. These 7 activities consist of a total of 689 samples, including climbing (102 samples), drinking water (96 samples), getting up from bed (101 samples), pouring water (100 samples), sitting down (96 samples), standing up (95 samples), and walking (99 samples).
    \item ASTRI Motion Dataset: This dataset involves activities such as walking, sitting, standing, squatting, and lying down performed by 11 participants, representing a diverse range of ages and genders. The data in this collection was captured using a single accelerometer integrated into a smart wristband, which could be worn on either the left or right hand. The accelerometer had a sampling rate of 52 Hz. The dataset consists of a total of 1080 samples, including walking (321 samples), standing (191 samples), squatting (189 samples), sitting (193 samples), and lying (187 samples).
\end{itemize}

\subsection{Baselines Algorithms}
We implemented various benchmarking activity recognition algorithms using wearable accelerometer sensor signals. These algorithms include Random Forest (RF) \cite{rf}, Support Vector Machines (SVM) \cite{svm}, Convolutional Neural Network (CNN) \cite{cnn}, Dynamic Time Warping (DTW) + 1 Dimensional CNN \cite{dtw1cnn}, DTW + Clustering \cite{dtwcluster}, Long Short Term Memory (LSTM) + Fully Connected Neural (FCN) network \cite{lstmfcn}, Temporal RP + ResNet (TRP+ResNet) \cite{reference1}, and modified Temporal RP + ResNet (MTRP+ResNet) \cite{reference1} algorithms.

To assess the individual contributions of our proposed method, we also implemented the Frequency domain RP plot with ResNet architecture (FRP+ResNet), as well as the Mixup Augmentation plot of temporal and frequency (Our Method). By including these additional variations, we aim to analyze the specific impact of different components in our approach and compare their performance against the baseline algorithms.

\begin{table}[h]
\addtolength{\tabcolsep}{-4pt}
  \caption{Baseline algorithms' performance comparisons with our method on ADL Dataset}
  \label{tab:adl_accuracy}
 \centering
 \begin{tabular}{|p{.9cm}|p{0.7cm}|p{0.65cm}|p{0.65cm}|p{0.65cm}|p{0.65cm}|p{0.65cm}|p{0.65cm}|p{1.1cm} |}
    \hline
    Method &  Climb stair & Drink glass & Getup bed & Pour water & Sit down & Stand up & Walk & Overall\\ 
    \hline
    RF & 75.3 & 90.8 & 76.8 & 88.7 & 87.2 & 88.2 & 89.9 & 85.2$\pm$2.7\\
    \hline
    SVM &  80.6 & 92 & 77.4& 86.9& 95.7& 96& 91.2& 88.4$\pm$2.4\\
    \hline
    CNN &  76.5& 83.3& 67.3& 83& 51& 76.8& 70.7& 72.2$\pm$1.8\\
    \hline
    DTW+ 1NN &  77.1& 94.3& 65.7& 92.9& 89.7& 93.7& 71.3& 83.3$\pm$3.5\\
    \hline
    DTW+ Cluster& 75.3& 89.4& 69& 89.5& 96.9& 95.5& 76.8& 84.4$\pm$2.8\\
    \hline
    LSTM+ FCN & 80.6& 86.8& 72.2& 95.7& 90.6& 92.2& 82.6& 85.7$\pm$2.0\\
    \hline
    TRP+ ResNet& 79.2& 83.3& 64.7& 83.6& 84.4& 78.5& 81.5& 79.3$\pm$1.4\\
    \hline
    MTRP+ ResNet& 82.8& 93.8& 81.2& 93.9& 97.8& 91.6& {\bf 94.3}& 90.9$\pm$1.6\\
    \hline
    FRP+ ResNet& 84& 85.2& {\bf 100}& 85.7& 70& 64.7& 66.7& 79.5$\pm$2.8\\
    \hline
    Our Method& {\bf 93.3}& {\bf 100} & 88.5& {\bf 97.8}& {\bf 100}& {\bf 100}& 88.5& {\bf 95.3}$\pm$1.8\\
    \hline
  \end{tabular}
\end{table}

\begin{table}[h]
\addtolength{\tabcolsep}{-4pt}
  \caption{Baseline algorithms' performance comparisons with our method on ASTRI Motion Dataset}
  \label{tab:astri_accuracy}
 \centering
 \begin{tabular}{|p{1.3cm}|p{0.9cm}|p{0.9cm}|p{0.9cm}|p{0.9cm}|p{0.9cm}|p{1.4cm} |}
    \hline
    Method &  Walking & Sitting & Standing & Squatting & Lying & Overall\\
    \hline
    RF & 88.7& 92.3& 88.7& 93.8& 86.5& 89.9$\pm$1.9\\
    \hline
    SVM &  93.4& 91.2& 90.8& 92.5& 87.2& 91.5$\pm$2.1\\
    \hline
    CNN &  85.2& 87.8& 81.3& 77.6& 71.4& 81.2$\pm$2.5\\
    \hline
    DTW+ 1NN &  53.3& 84.5& 93.4& 91.8& 57.4& 73.4$\pm$4.0\\
    \hline
    DTW+ Cluster& 65.7& 88.3& 90.4& 85.6& 70.2& 78.3$\pm$3.4\\
    \hline
    LSTM+ FCN & 92.4& 82.7& 88.9& 90.5& 82.3& 87.9$\pm$1.9\\
    \hline
    TRP+ ResNet& 91& 96.8& 79.3& 81.6& 79.1& 86.2$\pm$2.3\\
    \hline
    MTRP+ ResNet& {\bf 95.1}& {\bf 93.4}& 92.9& 98.3& 88.9& 93.9$\pm$2.0\\
    \hline
    FRP+ ResNet& 83.5& 88.7& 93.9& 100& 87.8& 90.78$\pm$3.4\\
    \hline
    Our Method& 84.1& 93.1& {\bf 96.4}& {\bf 100}& {\bf 100}& {\bf 94.72}$\pm$2.0\\
    \hline
  \end{tabular}
\end{table}

\subsection{Results Analysis}
We implemented both baseline algorithms and our proposed methods using various tools, including scikit-learn, libsvm, and TensorFlow in Python. To evaluate the performance, we used accuracy as the evaluation metric, calculated as the ratio of true positive and true negative predictions to the total number of predictions (accuracy = (TP + TN)/(TP + TN + FP + FN)). Additionally, we utilized the standard error, denoted by the plus-minus sign (±), as a measure of the distribution of errors. The standard error is calculated by dividing the standard deviation by the square root of the sample size. To calculate the accuracy, we followed a user-mixed approach that involved combining all episodes (data and labels), splitting the data into a 70:30 ratio for training and testing datasets, and performing training on the training data with 20\% random data selected as validation data during each iteration of neural network training. Finally, we evaluated the performance on the test data.

Table \ref{tab:adl_accuracy} and Table \ref{tab:astri_accuracy} presents the details accuracy comparisons of our proposed method with different baseline algorithms. Here, the central comparisons can be observed among three major versions of our proposed frameworks, modified temporal RP (MTRP), Frequency domain RP (FRP) and mixup augmentation of MTRP and FRP with ResNet. We can observe that, for ADL dataset (Table \ref{tab:adl_accuracy}), our proposed method outperforms all baseline algorithms in overall. However, only `Get up bed' activity performs highest with FRP and `Walk' performs highest with MTRP algorithms. On the other hand, for ASTRI Motion Dataset (Table \ref{tab:astri_accuracy}), our proposed mixup temporal and frequency domain RP image augmentation outperforms all baseline algorithms in overall while MTRP outperforms for walking and sitting detection.

Table \ref{tab:adl_accuracy} and Table \ref{tab:astri_accuracy} provide detailed accuracy comparisons between our proposed method and different baseline algorithms. These tables highlight the comparisons made among three significant versions of our proposed frameworks: modified temporal RP (MTRP), Frequency domain RP (FRP), and mixup augmentation of MTRP and FRP with ResNet.

When examining the ADL dataset (Table \ref{tab:adl_accuracy}), it becomes evident that our proposed method achieves superior performance compared to all baseline algorithms overall. However, it is worth noting that the activity 'Get up bed' achieves the highest accuracy with the FRP method, while the 'Walk' activity achieves the highest accuracy with the MTRP algorithm.

On the other hand, for the ASTRI Motion Dataset (Table \ref{tab:astri_accuracy}), our proposed mixup augmentation of temporal and frequency domain RP images surpasses all baseline algorithms in terms of overall accuracy. Furthermore, the MTRP method outperforms the others specifically for walking and sitting detection tasks.

\subsection{Conclusion and Limitations}
This paper presents a pioneering attempt to convert time-series data into recurrent plot images in the frequency domain. It demonstrates that while the frequency domain image representation alone may not always provide the most informative results, combining it with the temporal domain recurrent plot image representation surpasses existing methods with the help of advanced pre-trained image recognition models like ResNet. These novel findings open up new possibilities for time-series signal processing and its adaptation with scalable deep neural network models based on image processing. However, it is important to note that our proposed model was validated only on wearable accelerometer sensor signals. To establish the effectiveness of the temporal and frequency domain image representation technique, further validation is required on various time-series data such as Electroencephalogram (EEG), Electrodermal Activity (EDA), Photoplethysmograph (PPG), Gyroscope, Magnetometer, and others. Additionally, while our proposed method was evaluated solely on classification problems, it is necessary to validate it with other scalable machine learning techniques such as active learning, opportunistic learning, transfer learning, and reinforcement learning. Our long-term goal with this paper is to develop Automatic Scalable Machine Learning (AutoScaleML), which can represent any time-series signal using appropriate image representations that combine both temporal and frequency domain information, enabling the utilization of scalable computer vision models.

\end{document}